# The Definition of AI in Terms of Multi Agent Systems

Dimiter Dobrev
Institute of Mathematics and Informatics
Bulgarian Academy of Sciences
"Acad. G. Bonchev" Str., Bl. 8,
1113 Sofia, Bulgaria
dobrev@2-box.com

**Abstract:**

The questions which we will consider here are 'What is AI?' and 'How can we make AI?'. Here we will present the definition of AI in terms of multi-agent systems. This means that here you will not find a new answer to the question 'What is AI?', but an old answer in a new form.

This new form of the definition of AI is of interest for the theory of multi-agent systems because it gives us better understanding of this theory. More important is that this work will help us answer the second question. We want to make a program which is capable of constructing a model of its environment. Every multi-agent model is equivalent to a single-agent model but multi-agent models are more natural and accordingly more easily discoverable.

## Introduction

Some of the previous articles of the same author [1-13] address the issue of defining AI and the problem there is formulated in the same way, as it is in the articles concerning the multi-agent systems [14-16]. One may say that the formalisation used in [1-13] is a special case of the formalisation used in [14-16]. A special case results when we limit ourselves to a single agent.

Indeed, the formalisation in [1-13] is obtained as a special case from the formalisation of [14-16], but the articles [1-13] are nonetheless of interest, since they address a different issue. In other words, if we have two problems of the type 'given that…, find…', what is given will be the same in both problems, but what has to be found will be different. Still, those problems are significantly interconnected and this connection can be used both ways. That is, ideas from the second problem can be used to develop the first one and vice versa.

# Formulation of the problem with multi-agent systems

We are given one world, with multiple internal states, one of which is the initial state of the world. We have an *n* number of agents inhabiting this world. We also have one *World* function (this is the function of the world). It has two arguments. The first one is the current state of the world, while the second is the *n*-th set the actions of the *n* number of agents. The function of the world returns the new internal state of the world, which results from the current one and from the actions of the *n* agents.

In other words, if $s_0$ is the initial state, then $s_{i+1}=\text{World}(s_i, <a_1(i), ... , a_n(i)>)$, where $a_j(i)$ is the action of the на *j*-th agent at the *i* moment.

The question that is examined is whether a given agent has a strategy for achieving or maintaining a certain condition. This is summarized naturally by examining the same question with regards to a group of several agents. Such a group is called a coalition.

A strategy is the function which determines the actions of the agent (or of the coalition). What does this function depend upon? The only thing the strategy depends upon is $s_i$ (the current state of the world). The strategy is not dependent on 'History'. In other words, it is not dependent on the way the current state of the world has been reached. This means that it depends neither upon the $s_0, ... s_{i-1}$ states, nor upon the previous actions $a_1(t), ... , a_n(t)$ where $t<i$.

Just as it is not dependent on 'History', so isn't the strategy dependent upon 'Future'. This is only natural, since if our strategy were dependent on the future actions of our opponents, it would be rendered unusable, as the future actions of our opponents are unknown.

The last component that could influence the strategy is the *n*-th set of $<a_1(i), ... , a_n(i)>$. That is, the strategy may depend upon the actions of the opponents at the current moment. Let us assume that the *n* agents act simultaneously and that none of them is aware of the actions performed by the other agents at the same moment. This assumption holds the only undetermined aspect of the problem thus formulated. Had we assumed that the agents would be acting consecutively, with each seeing the actions of those before him, we would have had a completely determined system. In this case we would have had the attribute that if an agent does not have a strategy for maintaining a certain attribute then the coalition of the rest of the agents will have a strategy for achieving the negation of this attribute. (The opposite direction of this implication holds in both cases).

In order to see that with the current formulation the above attribute is not valid, let us consider the following example. Let us have a world with two states, *odd* and *even*. Let this world be inhabited by two agents, capable of two actions, 0 and 1, and let this world go into the *even* state when the sum under module two of the actions of the two agents is zero (the new state of the world does not depend upon the old one). In this situation, neither of the agents has a strategy for maintaining the *even* state or a strategy for achieving the *odd* state, i.e. the above characteristic is not valid.

## Formulation of the problem under the definition of AI

Here the formulation of the problem is the same, the difference being that the world is inhabited by a single agent (the AI in question). Another difference is that in the former case we assumed that the agents could see everything (i.e., they could see the current state of the world), while in the latter case it is assumed that the agent (i.e., the AI) receives only some of the information. In other words, now we assume one function, *View*, which restricts the information that the AI receives. Thus, the AI is like a horse with blinders, since it does not receive $s_i$ but only View($s_i$). In the isolated case when *View* is an injection function, the AI sees everything, but this is only possible in very simple worlds.

With the AI problem, the strategies cannot depend upon the current state of the world, simply because this state is not visible. That is why here they are dependent upon the 'Visible History' (this is the row View($s_0$), $d_1$, View($s_1$), ... , $d_i$, View($s_i$), where $d_i$ is the action of the AI at the moment $i$). In other words, the AI receives much less information than the agents from the first problem. Here the AI knows only what it saw and did, while in the first problem the agents know everything, since they can perceive the current state of the world, i.e., they see everything.

In the first problem it is assumed that we are given a concrete world and a concrete condition which has to be achieved or maintained. The question is whether there is a strategy that can achieve (respectively, maintain) this condition. In the problem with the AI, the world is unknown and the AI's task is to perform well in this unknown world. In order to define 'to perform well', we will need to introduce a meaning of life. To make our task easier, let us choose one particular meaning of life, as follows. Let there be two subsets in the set of values of the *View* function, which we will call *victory* and *loss*. The aim of life will be more *victories* and fewer *losses*. Of course, it is assumed that the AI distinguishes between those two subsets, i.e., it knows when it has won and when it has lost.

In the first problem the question is whether a strategy exists, while in the latter a certain, specific strategy is wanted. Or, more precisely, an AI is wanted, which is a computable strategy (we stated in [1] that the AI is a program, thus, it cannot be an incomputable strategy). It was proven in [6-7] that this strategy (i.e., the AI) exists; moreover, an algorithm was given, that calculates it. Unfortunately, this algorithm is completely useless, as it does not work within a reasonable timeframe. While an algorithm exists, which can complete the task with a finite number of steps, it becomes useless, when this finite number of steps is in all actuality infinite.

## How to elaborate the problem of the multi-agent systems

It is reasonable to assume that the agents cannot perceive everything but only a part of the world. In other words, we will assume that we have *n* functions View$_1$, ... , View$_n$, which limit the incoming information for the agents. In this case the strategy of the *j* agent will not be dependent upon the current state of the world but upon its visible history (which is the row $View_j(s_0)$, $a_j(1)$, $View_j(s_1)$, ... , $a_j(i)$, $View_j(s_i)$, where $a_j(i)$ is the action of the *j*-th agent at the *i* moment). I.e., the strategy can depend only upon what the agent has seen and done.

## How to elaborate the problem of the AI

The way of solving the problem of the AI goes through understanding the world. In other words, the AI must find a model of the unknown world in which it has been placed. (The algorithm from [6,7] does not function in this way, but as has already been said, this algorithm is not usable). If we are to search for a model of an unknown world, we must first choose a set of formal models, among which we will be looking for the one that will suit us best. Until now we were limited within the set of single-agent worlds but it is more practical to look for a formal model within the much larger set of multi-agent worlds. For each multi-agent world there exists an equivalent, single-agent one. In other words, by changing the set of formal models we are not changing the definition of the AI as stated in [1]. Still, the multi-agent worlds are more natural and it will be much easier to find a suitable model within their set. As was already said, for each multi-agent world there exists a corresponding single-agent one, which is usually much more complex and more difficult to comprehend than its multi-agent counterpart.

Why are the multi-agent worlds simpler? Because with them we can separate a part from the complexity of the world into a single object, which we will call an agent. Human beings think in much the same way. When we explore the world that surrounds us, we come across other people, whom we might take to be a part of the environment or, in other words, we might assume that we live in a single-agent world and that we are the sole inhabitants of the entire universe. Instead, we do accept that there are other people besides us and this makes the model of the world that surrounds us simpler and easier to comprehend. Often, we also accept the existence of more abstract objects, such as God, fate, luck. If you are to lock a person in a tower, thus placing them in a single-agent world, they are bound to start introducing additional agents into it, agents, which they do not actually need, in order to explain the surrounding world. For instance, such a person can assume that the wind is a rational being trying to tell them something. Or they can decide that a certain object is looking at them askance and may break it. In other words, people are created with a disposition to interpret the world that surrounds them as a multi-agent system and are, at times, apt to go somewhat too far in this direction.

When talking about multi-agent models, we should by all means make it clear that the different agents should be considered as equal. It is quite natural that the AI should divide the agents into *friends* and *enemies*. We can have a well-meaning, but stupid agent, which cannot be of much use. Similarly, another agent may be an *enemy*, and a cunning one at that, which will render it even more harmful.

When separating a part of the world to a particular agent, we will be adhering to the principle that the objects suitable for being transformed into agents, are those, whose behaviour is complex and difficult to anticipate. Usually, those are thinking objects, such as people or animals. Is there a point in regarding an object, such as a vacuum cleaner, for example, as a separate agent? The answer is 'not much', as the behaviour of the vacuum cleaner is quite clear, there is no hidden complexity to it which might be separated from the description of the world.

An important attribute of the AI is the ability to look at the world through the eyes of another agent, which is one more reason why the model of the world must be a multi-agent one.

## Note 1

Here is a simple proof of the fact that for each multi-agent world there exists a corresponding single-agent one.

**Proof.** Let us have a program which calculates the function $World_1$ of the multi-agent world and $n\text{-}1$ programs calculating the behaviour of the rest of the agents (i.e., all agents, except for the AI). Then the function $World_2$ of the respective single-agent world will be a composite of the above functions.

This proof poses three questions.

Firstly, whether there exist any functions describing the behaviour of the agents. The answer is 'Yes', since each agent is bound to perform a certain strategy, i.e., there exists a function describing its behaviour. In natural languages, what is meant by strategy is clever behaviour but here a strategy can be any behaviour.

The second question is whether the functions describing the world and the strategies of the agents are determined or not. We examined this question in [2] and saw that it is not of any real importance. Let us assume that the more common of the cases is present and that these functions are not determined.

The third and last question is why we assume these functions to be computable (i.e., that we have programs that can calculate them)? On one hand, this assumption is unnecessary, as the proof can do without it. On the other hand, there is nothing to stop us from assuming that these functions are computable, since we only have a finite part of their values (the part from the moment of birth to the present moment). In other words, we must approximate these functions using a part of them that is finite and there is nothing to stop us from choosing that the function for this approximation be computable.

How can a program calculate an undetermined function? Let us assume that these programs are using an inbuilt random-number generator. In other words, we will be departing slightly from the traditional concept of computability but if we assume the pseudo-random numbers to be random then the concept of computability will be same. In our case there is no need to distinguish between random and pseudo-random numbers, as we are dealing with a short interval of time (from the moment of birth to the present moment). I.e., the pseudo-random numbers are unlikely to start repeating themselves cyclically, nor are we likely to discover the laws that generate them.

**A concrete example**

In [4, 5, 11, 12] we discuss a concrete example – an artificial world (by 'artificial' we mean 'created by people'), where the AI must compete with an artificial opponent playing with it a game of tic-tac-toe. These articles attempt to find a single-agent model for this world – a rather difficult task, as the opponent is also part of the world, which renders the model quite complex. Of course, we have seen to it that the artificial opponent is well-defined, so that the world, too, is well-defined. Any indefiniteness of the opponent would have been transferred onto the world, as well.

In short, with a single-agent model the *World* function becomes too complex because its description contains the description of the artificial opponent. In such a world the artificial opponent is part of the landscape, which renders it rather complex, because whenever the AI places an X, an O appears on the board, only one O and only if the game is not over. It will be much simpler to assume not that the O's appear of themselves, but are being placed by an agent. We do not need to know the exact algorithm governing this agent (still, knowing it may only be of use). It should suffice to assume that this is a malevolent agent trying to harm us. It is exactly the idea of a malevolent agent that the *Min-Max* algorithm is grounded in (the algorithm by means of which chess programs 'think').

**Note 2**

When playing tic-tac-toe, we can choose to think that that the O's appear of themselves or that someone else is placing them. In much the same way, when sunbathing on the beach, we can choose to believe that clouds appear of themselves or that they are placed in the sky by somebody. In the former case we can call this 'someone' an 'opponent' and in the later – 'the god of suntan'.

If we assume that the O's (respectively, the clouds) appear of themselves, we might also assume that this is a completely random process, which we can neither foresee nor influence. This is just an assumption, but it is no good, as we do need to influence the process, so that we can win the game (or, respectively, get the tan we want). That is why we have to look for some complex interdependence, which might describe the appearance of the O's (or the clouds).

Assuming the presence of an opponent (or, respectively, of a god of suntan) will result in a simpler model, in which case it is only natural that we should try to influence the process by cheating our opponent or by propitiating the god of suntan.

A typical example of human behaviour is bringing gifts to a god or bribing a clerk with the aim of solving a certain problem. This strategy renders the question whether the model is adequate, for instance, whether the god or the clerk in question actually exist. A definite answer, be it 'yes' or 'no', is not possible in this case, as we do not normally communicate directly with gods or clerks, but have to resort to intermediaries instead. For example, it is the priest of suntan who receives the gifts and has to pass them on to the god. We cannot be sure that this god actually exists; we cannot even know for certain whether the priest exists or whether he is a figment of our imagination. I.e., seeing and hearing someone does not necessarily mean they exist, but we accept their existence, as the simplest explanation of what we see and hear.

But what is the answer to the question if the god of suntan exists? The answer is that it does not matter. What matters to us is whether by bringing gifts to this god we will ensure more sunny days. In other words, what matters to us is not whether a certain model is true but whether it works. And is it at all possible to talk about true and untrue models? If we divide the models into different classes, according to the equivalent relation, then each model has an infinite number, equivalent to it. Of course, we may consider the equivalent models undistinguishable and search for any representative of this equivalence class, but this is not going to solve our problem, either, as we are looking for the model of the basis of the visible history, but there is an infinite number of models which have the same visible history and yet are not equivalent. I.e., they differ in their future or somewhere else (in the alternative variants of the past).

In order to predict the future and plan our actions accordingly, we will have to choose from among the infinite number of possibilities a single, concrete model. But which one should it be? Normally, it is assumed that the simpler models are the more probable ones (the Ockham's Razor principle). Sometimes it is not easy to tell which is the simplest and most realistic model, which is why we just have to choose one and accept it as truth. For instance, we can accept the existence of the god of suntan and offer him gifts. This will hardly help, but most probably will not do any harm, either.

## Formalisation of the world from the example

In [4] we examine a formalisation of the world where the AI is playing tic-tac-toe. We have a board (3x3) and the AI has an eye, which it can move across the board. What is characteristic of this world is that the AI can only see a single box, while the rest of the board remains hidden and it must manage to imagine it. In other words, this is an interesting world because the AI cannot perceive everything in it (i.e., the *View* function is not an inaction). There are six possible moves at the AI's disposal – 'up', 'down', 'left', 'right', 'place an X in the current box' and 'new game'. The AI makes a new move by placing an X. When the game is over, the AI must clear the board, so that a new game can be started, which is why the command 'new game' is needed.

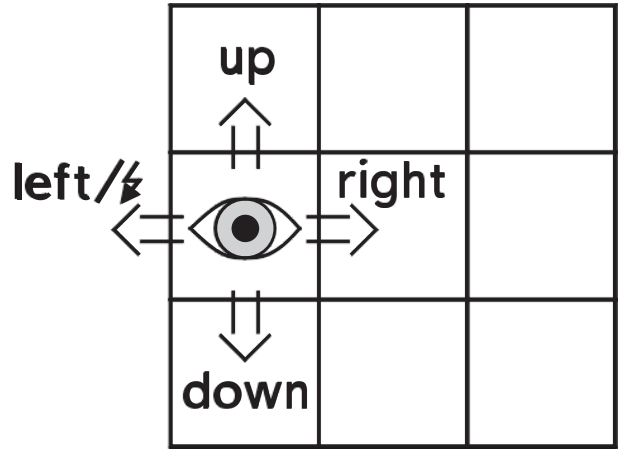

Figure 1 gives an idea of the moves the AI has at its disposal.

Up till now we described the possible exit of the AI, what remains now is to describe its entrance. The *View* function returns to us five bits of information. Two of them show the contents of the current box – 'X', 'O' or 'empty'. The other three bits are special. Let us call them 'victory', 'loss' and 'wrong move'. The first two provide us with a meaning of life (as we said, the meaning is more victories and fewer losses). The third bit is a guiding one and can be said to correspond to the pain instinct. Theoretically, the AI knows that wrong moves must be avoided but may at times choose to place one such move if it appears useful.

What about people? With people, the first two bits are missing, as the meaning of life is not defined, so they are left only with guiding instincts, such as pain and pleasure. In other words, if a person joins a game of tic-tac-toe, they may decide that the game is not that important to them and that it is not the meaning of their life, while for the AI there is nothing more important in the world than winning the game. At first glance it would appear that we have defined AI as something different from natural intellect, as the former has a clearly defined life goal and the latter does not. In truth, with our definition natural intellect is a special case of artificial intelligence, because we can limit ourselves to worlds where there is no victory or loss and where the AI, just as human beings, will have no clear goal.

## Multi-agent formalisation of the above world

As stated before, the multi-agent formalisation is useful, because we expect the resulting model to be simpler and more natural than a single-agent formalisation. In order to formalise the above world we can simply pick the artificial opponent and separate it into a separate agent. Of course, until now the artificial agent was well-defined (i.e., it played in accordance with a strictly assigned algorithm). In order to regard the artificial opponent as an independent agent, we will have to assume that it is not bound to a fixed playing algorithm but that it is instead capable of acting however it wants.

In this way we can easily move from a single-agent to a multi-agent model of the game. Still, this is not how we are going to do it, as this would result in an asymmetrical model and we would rather have the two agents equal.

**Why do we prefer symmetrical models?**

One of the considerable difficulties of AI is the necessity of providing it with education (the same, of course, holds true for natural intellect, too). We can provide a human teacher (a coach, a sparring partner, etc.) but this would be expensive and time-consuming, which is why we prefer the use of another program. In order to avoid having to write such a program specifically for this purpose, we prefer to let the AI play against itself, thus teaching itself.

Of course, in order for the AI to play against itself, the world does not have to be symmetrical, as the AI may assume different roles, so it is not a problem to use the AI for the different agents. The problem is that we want something more – an intellect governing a number of agents at the same time, much in the same vein as the idea of a computer governing a number of virtual machines.

If we have two agents in a symmetrical world, we can safely assume that they are governed by a single AI. This common intellect will discover the laws of the world at the same time for both agents but will maintain a different current state for each agent. For example, in the tic-tac-toe world, there is a very important finite state automaton providing information about the exact location of the eye of the AI on the board (*cf.* [4]). A lot of processor time is needed in order to discover this automaton, as it has to be sought among the set of all finite automata. That is why it is quite sensible to find it only once and then use it for both agents. Of course, as the eyes of the two agents will be in different columns on the board, this automaton will be in a different state with each of the agents.

As in our case we only have two agents, the advantages of a symmetrical world appear to be rather insignificant – we will need twice as little computer time, which is not of major importance. Still, when creating an AI, we will have to go through this stage of setting up and debugging. As was stated before, the AI is a program and the creation of any program must go through these stages, so we will find it much easier if both our agents are governed by the same program working in the same conditions. Otherwise we will end up with two AI's responsible for different agents and the resulting behaviours will be totally different as if the agents inhabited two different worlds. This is only too natural – if we feed different data to a program, its behaviour will inevitably be different.

There are other cases, where the advantages of a symmetrical model are quite significant, namely, when we have thousands of agents. For instance, let us imagine how the first robots to appear in a shop will function. We may assume that they will all be autonomous, each with its own AI (i.e., its own computer, performing a separate copy of the AI program). This assumption is rather doubtful, not least because it would require that each robot be taught separately. A much more probable assumption is that the robots will have small computers, compressing the input information and transferring it to a central computer which will execute a copy of the AI program. This central AI will have a number of virtual agents (our robot being one of them) and will be controlling them directly.

Such a model will have a number of advantages. For one, we will not have to teach each and every robot how to prepare a lasagne – after we have taught one of them and all the rest will know how to do it. We will not have to introduce ourselves to each robot separately – once one of them knows us, they all will. We could even use the robots as telephones. We could just say "I want Pete's robot to tell Pete that he is invited over tonight". Of course, our robot can use the phone to call Pete's robot, but that will not be necessary, since both robots are governed by the same brain. To paraphrase the saying, 'the left hand cannot but know what the right one is doing'.

The advantages of such a model become even more obvious if we examine the scheme of street traffic. What we have today is a flow of vehicle operated by people each with their own autonomous intellect. In order to synchronise this flow, we need traffic lights, street marking, traffic codes, etc., the result being a flow of traffic that is both slow and unsafe.

Let us now imagine a flow of vehicles operated by AI and by a single, shared AI, at that, as opposed to each car having its own 'opinion'. How might a road junction look then? Cars will be driving at great speed along both streets, passing each other without any traffic lights and yet, without crashing into each other. How will this be possible? The secret is that all traffic participants will be governed by a single intelligence. When there is only one person on the dancing floor, they have no problem synchronising the movements of their left and their right hand, but when there are two dancers, the problem of synchronisation becomes almost unsolvable. With dancing there are a lot tricks similar to the ones used in street traffic, tricks such as learning all movements beforehand or agreeing on having one partner lead and the other follow. A much better result might be achieved if the dancing couple were governed by a single intelligence – we would then have perfect synchronisation without any need for rules and beforehand agreements.

## How will the concrete example look like?

In our example of a symmetrical two-agent world we have one board (3x3), with each agent seeing only one box at a time and being capable of moving across the board independently of its opponent. The internal state of the world is determined by the position on the board, the coordinates of the eye of the first agent and the coordinates of the eye of the second one. There will be two more bits of information showing which agent's turn it is and whether the game is over (i.e., whether one of the agents is expected to place an X or to say 'new game'). In order for the agent to be able to place an X, it must be its turn. If it is not, but the agent tries to place an X, it will only receive the message 'wrong move' and no X will appear on the board.

As we said before, the game is tic-tac-toe. When one of the agents wins, it will see 'victory' and the other will see 'loss'. One of the agents will always place X's and the

other – O's. In order to have a symmetrical world, we will have to cover the eye of the second agent with a pair of 'magical glasses', through which it will see the X as an O and vice versa, so both agents will think they are placing X's and their opponent – O's. Still, the world will not be entirely symmetrical, since at the initial moment it will be one of the agents' turn, but not the other's. This is, however, a minor asymmetry and as such of no real importance.

We will also introduce a time-trouble rule, so that neither of the agents can block the game by moving across the board without placing a move. The time-trouble rule will state that each agent whose turn it is has only ten steps in order to place its move. If it fails to do so, it will lose the game and the turn will pass to the other agent (so that it may say 'new game', thus clearing the board). If neither agent does anything, after each ten moves one of them will lose a game (after another ten – the other will lose). The number ten is, of course, random; if we change it, we will have another world with different game rules, but it is wiser to decide upon a certain number, so that we can think within the terms of a concrete world.

Because of the time-trouble rule, we will add to the internal state of the world a counter, which will count the number of steps after the turn has passed to one of the agents. This counter will not be visible to the agents because their *View* function will not show it to them. Of course, the agent can count to ten, the problem is that it won't know when to start counting. When it sees a new O on the board, it will know that its opponent has placed its move and that it has ten steps in order to place his own, but it won't be able to tell how many steps it has left, as it won't know how many steps ago it was that its opponent placed this new O. The strategy of the player, whose turn it is not, will be to move across the empty boxes, looking for a new O, trying to place a move from time to time with the assumption that its opponent has already played and that it is now its turn.

We could also base the time-trouble rule not upon the number of steps taken but upon the amount of real time that has passed. This, however, would complicate the model, because all will depend upon the amount of time that the AI needs in order to make a step. It will be much easier to assume that this time is a constant, i.e., that a certain number of steps corresponds to a certain amount of real time. Or we can choose to think that real time does not exist and that the only thing we have is number of steps in an artificial world.

If we are to look at this two-agent world from the standpoint of the first agent, we will get a single-agent world, similar to the one, described in [4,5,11,12]. Where is the difference? In that world, when the AI placed an X, an O immediately appeared on the board, while here this happens after a certain number of steps. Earlier, the inbuilt opponent saw everything and was, accordingly, able to place its move at once. Now the other agent sees what the AI sees and needs time (i.e., steps) in order to go around the board and examine it.

## How to reach the *Min-Max* Procedure

In [11] we managed to build a model of this world (the world there was roughly the same). We managed to find a finite state automaton describing the current coordinates of the AI's eye. We also found a predicate describing the position of the board (a three-place predicate whose arguments were the *x* and *y* coordinates of the eye and the time, i.e., the number of the current step). On the basis of this predicate and of first-row formulae, we described the conditions for reaching a victory. In other words, in [11] we found a world model which correctly described the rules of the game. However, we could not finish our task and could not build a program capable of playing tic-tac-toe on the basis of this program.

**Note.** The predicate describing the position of the board returns three possible values, which is why it is better to talk not of a predicate but of a function or a pair of predicates, one of them describing the X's and the other – the O's. Those, however, are technical details, which we will overlook.

What was left unfinished in [11] was reaching the *Min-Max* procedure, or another, similar one, like the *Max-Sum* one (*cf.* [6]). It is exactly the idea of the multi-agent model that will help us apply the *Min-Max* procedure.

If we want to make the AI plan its future moves, it is only natural to apply *Min-Max* on the principle AI versus the world. In other words, we calculate the maximum of all possible steps that the AI could take and then the minimum of all possible reactions of the world. We discussed such a possibility in [9] and found that it leads directly to a combinatorial explosion. Please note that to each move there are about ten corresponding steps (because the AI and its opponent must first move to the empty box before they can place their move). This means that the tree of the *Min-Max* procedure becomes about ten times higher. And since we know that the complexity of *Min-Max* is exponentially dependent on the height of this tree, it becomes clear at once that this way leads directly to a combinatorial explosion.

To render the *Min-Max* procedure functioning, it must be based on the principle AI versus the other agent. To this aim the AI must realize that what matters most in the world of tic-tac-toe is the position of the board. Of course, this position is not directly visible, i.e., it is an abstraction. As we saw in [11], the AI can find the predicate describing the position on the board or, in other words, this is an abstraction that the AI can understand.

The next thing that the AI must realize is that it can change the position on the board. It is not difficult to deduce that placing an X in an empty box will lead to a change in the position (i.e., this will change the value of the predicate for the current *x* and *y* and the next *t*). It will not be difficult for the AI to realize that it can move across and thus reach any of the empty boxes. This means that the AI can recognise the possible moves it has at its disposal.

The next abstraction that the AI must reach is the idea of the virtual opponent inhabiting the same world and having the right to place an O in any of the empty boxes.

Now the AI is ready to apply the *Min-Max* procedure. The only thing left for it to decide is whether the other agent is friend or foe, or, in other words, whether the procedure should be *Min-Max* or *Max-Max*. This should be easy. Let the AI be a believer in good and assume that the other agent is a friend. This will very soon change, when, instead of helping like a friend, the other agent begins playing in the most hostile possible way. In other words, the other agent will quickly lose the trust of our AI and will be considered a foe.

**Variants of the world from the example**

Let us have a look at several variants of the above world in order to see that *Min-Max* isn't the only possibly strategy for the AI. In other words, the world may be of the 'me and a single foe' type, but there are other variants, as well.

Let us take the same world, but this time, inhabited by four agents playing tic-tac-toe. Let the first one place an X, the second one an O, etc., or, in other words, the first and the third agent will be playing against the second and the fourth one. The procedure in this case will be of the type *Max-Min-Max-Min*, where the first *Max* is of the possible moves of the AI and the second – of those of the third agent (the one playing on the side of the AI).

Let us now change the game and assume that the third agent has been bribed by the other team. I.e., it places an X, but does so in the worst possible way, instead of in the best. The procedure in this case will be *Max-Min-Min-Min*.

Let us now assume that agents three and four do not think but play completely randomly. This will result in a *Max-Min-Average-Average* procedure, where *Average* will stand for the arithmetic mean of all possible moves.

In all the variants that we looked at, the agents play consecutively (i.e., each waits for its turn). Let us now look at a variant of the example where the agents play simultaneously. Let us in the beginning of the game have the first agent have two X's, which it can place on the board any time it likes and let it get a new X after each ten steps. Let the same go for the second agent. There will be no order in this example (i.e., no waiting for turns), nor any time-trouble rules. In case both agents try to make a move in the same box, let the first one has priority. Should both of them make a line at the same time, the game will be considered a draw. It is still possible to use a procedure similar to *Min-Max* in worlds like this, even though the tree of the possible developments will be much more complex than in the world where the agents are waiting for their turns.

**Conclusion**

Multi-agent models are the right way to go when solving the problem of creating an AI. Even though the AI algorithm is to a great extent clear, there still remain to be solved a number of technical and conceptual problems, before a functioning prototype could be created.